\documentclass[preprint]{article}
\usepackage{neurips_2024}

\usepackage{xcolor}
\usepackage[utf8]{inputenc} 
\usepackage[T1]{fontenc}    
\usepackage{hyperref}       
\usepackage{url}            
\usepackage{booktabs}       
\usepackage{amsfonts}       
\usepackage{nicefrac}       
\usepackage{microtype}      
\usepackage{amsmath}
\usepackage{graphicx}
\usepackage{longtable}
\usepackage{multirow}
\usepackage{array}
\usepackage{longtable}

\usepackage{subcaption}
\newcolumntype{H}{>{\setbox0=\hbox\bgroup}c<{\egroup}@{}}

\title{Hopfield Networks for Asset Allocation}

\author{%
  Carlo Nicolini\\
  Ipazia SpA\\
  Milan, Italy\\
  \texttt{c.nicolini@ipazia.com} \\
  \And
  Monisha Gopalan\\
  Ipazia SpA\\
  Milan, Italy \\
  \texttt{m.gopalan@ipazia.com}\\
  \And
  Jacopo Staiano \\
  University of Trento\\
  Trento, Italy\\
  \texttt{jacopo.staiano@unitn.it}\\
  \And
  Bruno Lepri \\
  Fondazione Bruno Kessler\\
  Trento, Italy \\
  Ipazia SpA\\
  Milan, Italy\\
  \texttt{lepri@fbk.eu}
}

\begin{document}

\maketitle

\begin{abstract}
We present the first application of modern Hopfield networks to the problem of portfolio optimization. We performed an extensive study based on combinatorial purged cross-validation over several datasets and compared our results to both traditional and deep-learning-based methods for portfolio selection.
Compared to state-of-the-art deep-learning methods such as Long-Short Term Memory networks and Transformers, we find that the proposed approach performs on par or better, while providing faster training times and better stability.
Our results show that Modern Hopfield Networks represent a promising approach to portfolio optimization, allowing for an efficient, scalable, and robust solution for asset allocation, risk management, and dynamic rebalancing. 
\end{abstract}

\section{Introduction}
Starting from the seminal work of Markowitz~\cite{markowitz1952}, financial literature has seen multiple attempts to tackle some of the fundamental problems of portfolio optimization, like the estimation of covariance matrix and asset returns. 
Within the standard portfolio optimization theory, a significant challenge is posed by the accurate estimation of asset returns:
misestimations can lead to sub-optimal asset allocations, thus increasing the risk of financial losses and undermining the efficiency of the investment strategy.
While some authors adopt theoretically sound approaches (the ideas from~\cite{he2002intuition} or ~\cite{meucci2011robust, meucci2010fully} to cite a few), several tend to be guided by practice and intuition, such as the bootstrap-based resampled efficient frontier by \citet{michaud2007estimation}.

Recently, the work of \citet{zhang2020deep} sparked a wave of interest towards the application of deep learning to portfolio optimization: the authors formulated the portfolio optimization problem within an end-to-end approach based on maximizing the strategy's Sharpe ratio by means of a Long-Short-Term-Memory (LSTM) network~\citep{hochreiter1997long} to learn asset returns' temporal patterns.
The memory cells of the LSTM layer are good at modeling asset returns as they are able to generalize quite well to the test set, resulting in high out-of-sample Sharpe ratio.

A more complex architecture was proposed in the \textit{Portfolio Transformer} by \citet{kisiel2022portfolio}. 
Their model is largely inspired by the standard Transformer architecture proposed by \citet{vaswani2017attention}: a 4-layer encoder is connected to a 4-layer decoder in a very similar way to the original structure proposed for machine translation tasks.
Differently from the standard Transformer, where positional encoding is adopted to inject temporal dependencies to the self-attention layer, the authors of the Portfolio Transformer embed asset returns through a Time2Vec layer~\citep{kazemi2019time2vec}, in order to learn frequency and phase parameters within positional embeddings.

Inspired by these two works, and guided by a physical intuition~\cite{krotov2020large}, we adopt a neural architecture based on Modern Hopfield Networks~\citep{ramsauer2020hopfield}, here abbreviated as MHN.
This recently proposed architecture is the continuous counterpart of traditional Hopfield networks, with an exponential activation function leading to a much larger repertoire of stored patterns~\citep{demircigil2017model,krotov2016dense,krotov2023new}. 
The expressive power of MHNs, together with their fully connected, recurrent architecture~\cite{hopfield2007}, makes them suitable for tasks where time-series are provided in input.

At the core of our approach is the mitigation of the adverse impacts of estimation errors inherent in the Markowitz method~\citep{boyd2024markowitz,sexauer2024harry}, while simultaneously preserving the high-capacity storage and retrieval capabilities for intricate patterns, a distinctive feature of MHNs~\cite{ramsauer2020hopfield} that is reflected in better estimation of future returns.
By leveraging their high storage capacity, MHNs can assimilate a broader spectrum of market conditions and anomalies, leading to more robust and accurate return estimates in adverse regimes.

In this work, we
empirically test this hypothesis via an experimental setup including two MHN-based architectures: the first design simply replaces the LSTM layer in~\cite{zhang2020deep} with an Hopfield Pooling operator, internally guided by a dense associative memory mechanism; in the second approach, we use a Transformer-like architecture but with an Hopfield layer replacing the multi-head self-attention operator.

Our results reveal that a simple MHN often surpasses more intricate architectures such as LSTMs, achieving superior performance within significantly shorter training times, and thus underscoring the efficacy of neural architectures within the domain of portfolio optimization.

In the following we describe both models in detail, present the benchmark results and finally discuss implications and possible problems of our proposed approaches.

\section{Background and methods}
\subsection{Notation and metrics}
In this study, we analyze time series of \textit{daily} asset prices for the portfolio optimisation task. 
We denote the time series of asset prices as a $T \times N$ matrix $\mathbf{P}$ with elements $P_{t,i}$ denoting asset $i$ at time $t$.
Similarly, asset returns are denoted by a $(T-1) \times N$ matrix $\mathbf{R}$ with elements $R_{t,i}$.
We compute asset returns as log-returns $r_{ti} = \log P_{t,i} - \log P_{t-1,i}$ instead of simple returns, as they are leaner to work with and tend to be better distributed towards a Gaussian distribution~\citep{campbell1998econometrics}.

Asset weights in the portfolio are denoted by a vector $\mathbf{w}$ with shape $(1,N)$ and elements $0\leq w_i \leq 1$. 
For long-only portfolios asset weights are all positive and sum to 1. 
One could also deal with short portfolios, having positions with the only constraint of summing to $1$, thus allowing negative weights.
In this work, however we only discuss long portfolios.

The portfolio return series $R_p(t) = \mathbf{w}^T \mathbf{R}(t)$ allows for the computation of various performance metrics, such as the average return, Sharpe ratio, and Sortino ratio. 

The Sharpe ratio~\citep{sharpe1966mutual} evaluates risk-adjusted return by comparing the average portfolio return $\bar{R}_p = \langle R_p \rangle$ to its standard deviation (or volatility) $\sigma_p = \sqrt{\langle (\bar{R}_p - R)^2} \rangle$, relative to the risk-free rate $R_f$.
Meanwhile, the Sortino ratio~\citep{rollinger2013sortino,sortino1994performance} focuses on downside risk, using downside deviation $\sigma_d=\sqrt{(\min(\bar{R}_p-R,0)^2}$ to assess volatility based solely on negative returns, providing a nuanced view of portfolio performance under adverse conditions. 
These metrics are essential tools in modern portfolio analysis, as highlighted in Chen's comprehensive framework~\citep{chen2016modern}.

We also reported the Average Drawdown~\citep{chekhlov2004portfolio} (\texttt{Avg.DD}), computed as the average of all drawdown series over the portfolio history. It provides an average measure of the peak-to-trough decline in asset value over multiple drawdown periods, offering insights into the consistency and severity of losses experienced by an investment.

\subsection{Time2Vec embeddings}
The input of our models are the asset returns, with a first layer operating a transformation of returns from the time domain to the frequency domain.
Specifically, we employ Time2Vec~\citep{kazemi2019time2vec} to incorporate trainable embeddings into the model parameters. 
Each univariate time series representing the $i$-th asset returns $r_i(t)$ undergoes a transformation onto a $K+1$ dimensional vector $\mathbf{v}_i$ through learned frequencies and phase shifts $\omega_k$ $\phi_k$, hence capturing periodic rhythms within $r_i(t)$ through a periodic activation function $\mathcal{F}$ (like sine or cosine):

\begin{equation}\label{eq:time2vec}
v_i^k(t) = \left\{ \begin{aligned} 
  \omega_k t + \phi_k & \quad \textrm{if} \quad k=0\\
  \mathcal{F}\left(\omega_k t + \phi_i \right) & \quad \textrm{if} \quad  1 \leq k \leq K
\end{aligned} \right.
\end{equation}

Working with multivariate time series, we 
embed each asset time series consisting of $T$ observations within each batch and concatenate the embeddings.

\subsection{Hopfield Networks}
Modern Hopfield Networks~\cite{ramsauer2020hopfield}, like their traditional counterparts~\cite{hopfield1982neural} are densely connected Recurrent Neural Networks~\cite{sutskever2013training}. 
Differently from their traditional counterpart, they can work with continuous input, store an exponentially large number of patterns and are able to retrieve patterns with a single update~\cite{ramsauer2020hopfield}. 
To be precise, let $\mathbf{X} \in \mathbb{R}^{N \times d}$ be a matrix of $N$ data samples $\mathbf{x}_1,\ldots \mathbf{x}_n$ (also called \textit{memory patterns} in Hopfield networks literature), each one with dimension $d$.
Memory patterns $\mathbf{X}$, such as asset returns or their corresponding \textit{Time2Vec} embeddings, are forwarded through the Hopfield network. 
The backpropagation step iteratively adjusts the network's internal state pattern $\mathbf{q}_t \in \mathbb{R}^d$ converging towards stored patterns $\mathbf{X}$ or their combinations.
This process has an associated energy function described by Equation~\ref{eq:hopfield_ramsauer}, first described by \citet{ramsauer2020hopfield}:
\begin{align}\label{eq:hopfield_ramsauer}
\text{E} &= -\beta^{-1} \log \left( \sum_{l=1}^N \exp(\beta(\boldsymbol{X}^T \mathbf{q})_l)) \right) + \frac{1}{2} \mathbf{q}^T\mathbf{q} + \beta^{-1}\log N + \frac{1}{2} M^2.
\end{align}
The last term in the equation is the regularization term where $M$ is the largest norm of all stored patterns: $M = \text{max}_i ||\mathbf{x_i}||$.
The coefficient $\beta$ has its foundation in statistical physics and is often considered to act as an inverse temperature. In Equation~\ref{eq:hopfield_ramsauer} it acts as a regularization term, controlling the balance between the free-energy term (first addend) and the internal energy (second addend).

Interestingly, as pointed out by \citet{ramsauer2020hopfield} the update rule for internal states $\mathbf{q}$ defines exactly the attention mechanism used within the Transformer model: direct minimization of the energy-based model~\ref{eq:hopfield_ramsauer} yields an update rule 
\begin{equation}\label{eq:updatestep}
\mathbf{q}_{t+1} = \textrm{softmax}\left(\beta \mathbf{X} \mathbf{q}_t \right)
\end{equation}
In this sense, Modern Hopfield networks can be seen as the theoretical background behind the Transformers' attention mechanism.

\subsection{Model architecture}
\begin{figure*}[htb]
\begin{minipage}[t]{0.45\textwidth}
\includegraphics[width=\textwidth]{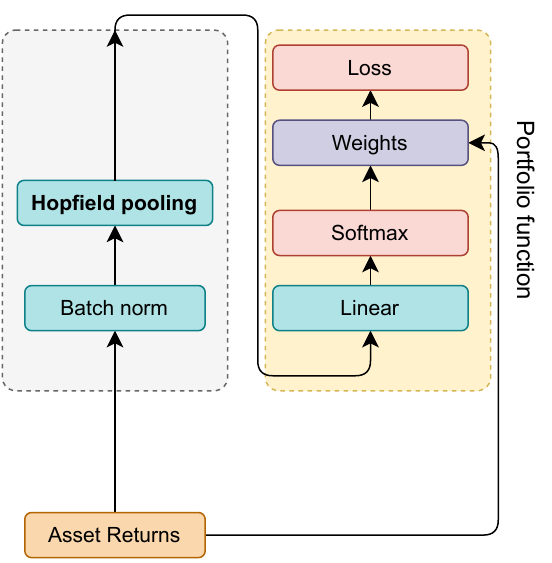}
\end{minipage}%
\begin{minipage}[t]{0.5\textwidth}
\includegraphics[width=\textwidth]{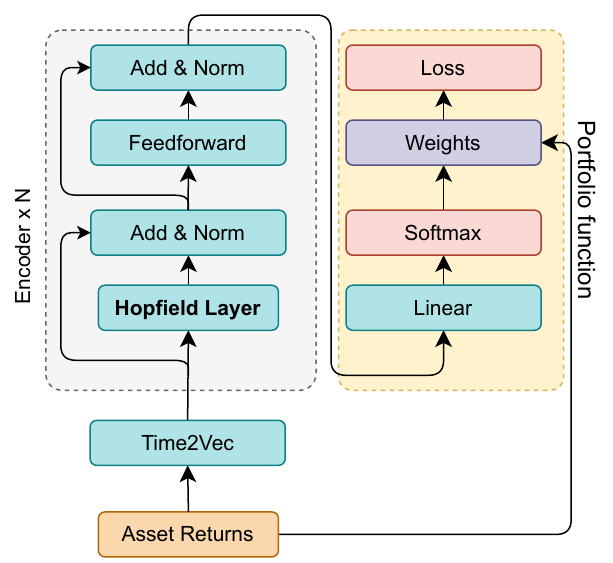}
\end{minipage}
\caption{\textbf{Left:} The Hopfield Pooling model (\texttt{HOP-POOL}), where a simple pooling operator replaces an LSTM layer. \textbf{Right:} In our Hopfield Encoder model (\texttt{HOP-TRA}) we stacked the encoder block four times (dashed gray blocks denoted by \textsf{Encoder x N}).}
\label{fig:hopfieldmodel}
\end{figure*}

We adopt the energy model operationally described by Equations~\ref{eq:hopfield_ramsauer} and \ref{eq:updatestep} within two effective network architectures: Pooling~\citep{gholamalinezhad2020pooling} and Transformer~\cite{vaswani2017attention}.
The first architecture, termed the Hopfield Pooling (Figure~\ref{fig:hopfieldmodel}, left panel), resembles the foundational architecture proposed by \citet{zhang2020deep}, albeit with the substitution of the LSTM component by a Hopfield Pooling layer.

The second architecture, denoted as the Hopfield Encoder (Figure~\ref{fig:hopfieldmodel}, right panel), draws inspiration from the Transformer's encoder, wherein the multi-head self-attention mechanism is substituted by a Hopfield layer.

\paragraph{Hopfield Pooling}
We build over the idea of \cite{zhang2020deep}, by replacing the original Long Short-Term Memory (LSTM) network~\cite{hochreiter1997long} with an Hopfield Pooling layer. 
The Hopfield Pooling layer is a neural operator mapping $[B,T,N]$ tensors into $[B,N]$ arrays, hence operating a pooling operation on the temporal dimension. 
The recursive properties of the Hopfield layer are used with a hidden neuron dimension of 2048, thus allowing to store long and complex patterns. 
More specifically, this large associative memory provided by the Hopfield Pooling layer allows to capture and recall complex patterns during down-sampling and feature extraction, hence propagating rich information to the downstream layers.

\paragraph{Hopfield Encoder}
Here, we maintain the encoder part of the Transformer, but replacing the multi-head self-attention mechanism with a Hopfield layer.
Following the Time2Vec embedding process, tensors denoted as $\mathbf{X}$ with a shape of $K+1$ are forwarded into the encoder.
The encoder comprises of replicated modules of a designated encoder block function denoted as $f$. 
The function $f$, as described in Equation~\ref{eq:encoder}, consists of a Hopfield layer, succeeded by a skip connection and layer normalization as per the formulation proposed by \citet{ba2016layer}. 
Subsequently, the output undergoes processing by a multilayer perceptron (MLP) featuring the GELU activation function proposed by \citet{hendrycks2016gaussian}. 
This MLP output is then subjected to an additional skip connection, as outlined by \citet{he2016deep}, followed by another layer normalization step.

In progressing from block $b$ to block $b+1$, each data tensor $\mathbf{X}$ undergoes two transformations as described in Equation~\ref{eq:encoder}:
\begin{align}\label{eq:encoder}
\mathbf{X}' \leftarrow \textrm{\textbf{LayerNorm}} \left ( \textrm{\textbf{HopfieldLayer}}(\mathbf{X}) + \mathbf{X} \right) \\
\mathbf{X}_{b+1} \leftarrow \textrm{\textbf{LayerNorm}} \left( \textrm{\textbf{MLP}}\left( \mathbf{X}' \right) + \mathbf{X}' \right) \nonumber
\end{align}

By stacking multiple layers of the encoder, the model progressively captures increasingly intricate representations of the structural properties inherent in the time series data.
The hidden size dimension of the MLP is set to four times the model embedding size.
For computational efficiency and to match the architecture by~\cite{kisiel2022portfolio}, we opted for an embedding dimension of $D=128$ and implemented a configuration comprising four layers in the encoder.

\paragraph{Output layer}
For both the pooling and the encoder models, the output tensor $\mathbf{X}'$ at the last block is fed through a final dense layer (\textit{Linear} in Figure~\ref{fig:hopfieldmodel}) reshaping the $D$-sized input onto $N$ shaped logits.
Portfolio asset weights then result from the $\textrm{softmax}$ operation applied on the output logits. 
In the simplest settings, the output layer consist of a softmax that normalizes the weights to unit-sum, followed by the calculation of the loss function.
Alternatively short portfolios are possible, by taking the output $s_i$ from the last layer \textit{(logits) $s_{it}$} of the model, considering the sign and normalizing through a softmax function over the assets~\citep{zhang2021universal}.

Our methods make it possible to implement any differentiable portfolio metric in terms of a loss function. We experimented both with maximization of strategy's Sharpe ratio $\bar{R}_P/\sigma_P$ and minimization of strategy volatility $\sigma_P$.

\subsection{Models training}
\begin{figure}[htb]
\centering
\includegraphics[width=0.49\textwidth]{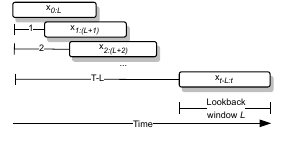}\hfill
\includegraphics[width=0.49\textwidth]{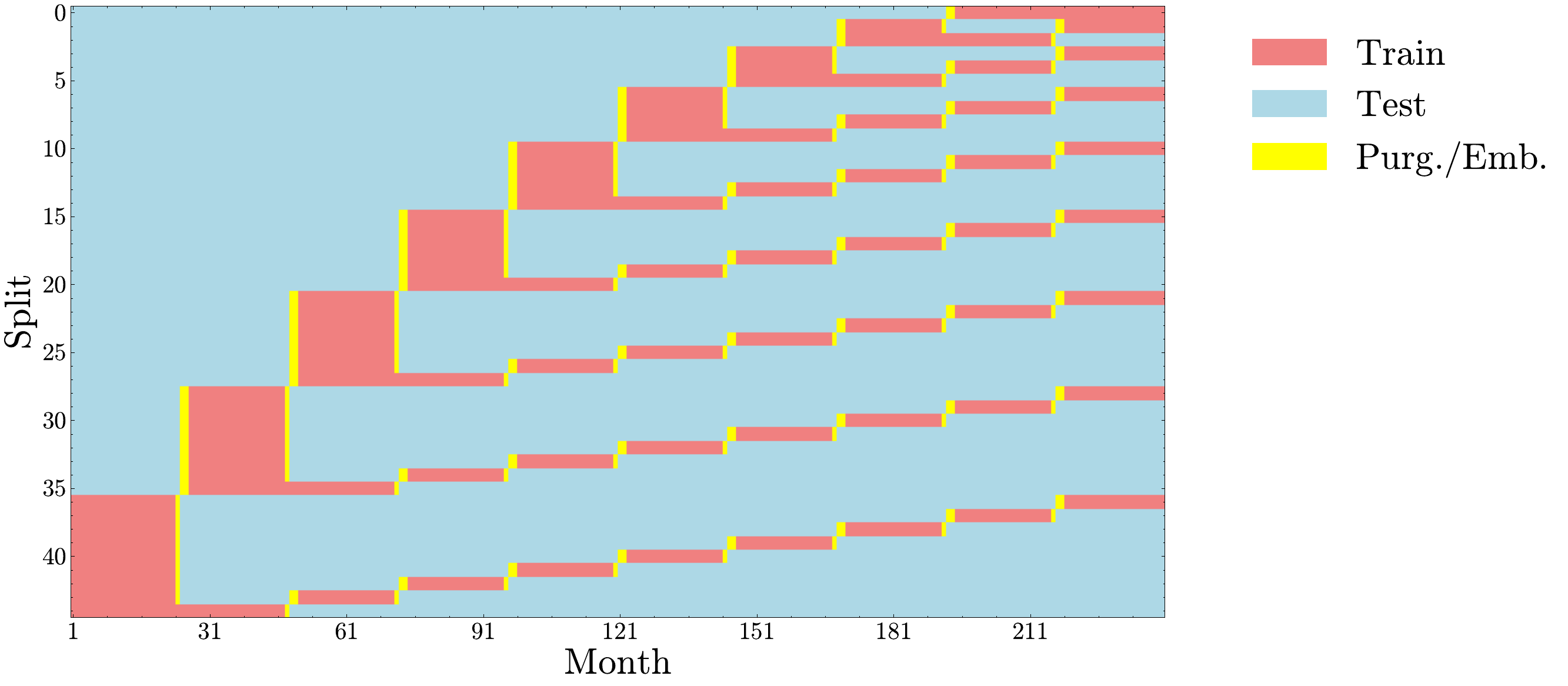}
\caption{\textbf{Left:} Data batching procedure. For each training and validation set, $B$ batches are created of shape $L,N$ for each batch, where $L$ is the look-back window and each batch element is shifted by one observation from the previous, until the final element is reached. \textbf{Right:} Schematic representation of training and testing folds in combinatorial purged cross-validation. For visual clarity we aggregated folds over 20 years of data at monthly granularity.}
\label{fig:batching-crossval}
\end{figure}

Training data are forwarded to the network in mini-batches of size $B=32$, with each batch $b\in \lbrack 1,B \rbrack$ containing observations from the training set with rows $\lbrack b+t, \ldots, b+t+t_l \rbrack$ where $t_l=128$ days is the look-back window.
This batching procedure is illustrated in the left panel of Figure~\ref{fig:batching-crossval}.
We apply the same batch size both for training and validation data. 
This data batching approach facilitates efficient learning from the time series sequential data.

At inference-time, portfolio weights are obtained as the average of the predicted weights over all input batches.
This marks a significant difference with the work of \citet{zhang2020deep}, whereas they implicitly assumed daily rebalancing within the (walk-forward) backtesting.

We have utilized AdamW for model training, an adaptive momentum strategy ~\citep{kingma2014adam,loshchilov2018fixing} that works best in terms of out-of-sample results' quality.
We have limited the number of epochs to 500, however we have found that over a multitude of datasets, typically the optimizer finds a good solution reaching good generalization around epoch 10-20.
Unsurprisingly, we have seen that overtraining the model, even in conditions of better validation loss, tends to produce bad test-time allocations when evaluated with a fair cross-validation technique.

For model training we have utilized an AWS \texttt{g5.2xlarge} with 8 CPUs and a single NVidia A10 GPU equipped with 24GB of VRAM. 
The Hopfield Encoder has the highest utilization of GPU resources and training on 20 years of daily data for almost 200 epochs takes around ten minutes.

\subsection{Cross validation and statistical testing}
The combinatorial purged cross-validation (CPCV) method introduced by \citet{deprado2018advances} was employed to assess the performance of the portfolio allocation strategy. 
The CPCV method tries to overcome the overfitting challenges faced by traditional k-fold and walk forward cross validation by generating multiple paths of combinations of training and test data. 
Moreover, in order to make our comparison more fair, purging was applied to remove any data leakage between the training and test sets in addition to embargoing to remove any serial correlation.

To minimize the probability of false discoveries, we have used 10 training folds and 8 test folds, resulting in 36 back-test testing paths and 45 training combinations.
The average training set size is $\approx$ 3.5 years, while the test size is $\approx 2$ years.
For both the purging and embargoing periods we have set 21 days, corresponding approximately to one business month. 

Within each cross-validation training fold we further subdivided the training set into two distinct consecutive sets in proportions of 80\% training and 20\% validation, in order to keep track of the loss function and trigger early stopping.
We have used the \texttt{skfolio}\footnote{\href{https://github.com/skfolio/skfolio}{https://github.com/skfolio/skfolio}} implementation of combinatorial purged cross-validation.

Figure~\ref{fig:batching-crossval} depicts the cross-validation setup we used.
\paragraph{Transaction costs}
Importantly, in order to focus on specific behaviours of our methods, in the following Results Section we have ignored transaction costs and management fees.

\subsection{Datasets}
We compare our method against traditional baselines over four different datasets representing different sectors (stocks, commodities, ETFs) hence having different returns' properties.
All the datasets cover a span of almost 20 years.

The first dataset (named \texttt{ETF} in the text) consists of the four ETFs used by \citet{zhang2020deep}, although over a different temporal interval. The period analyzed spans from February 6th 2006 to December 29th 2023. 
The four assets are \textit{iShares Core U.S. Aggregate Bond ETF} (\texttt{AGG}), CBOE Volatility Index (\texttt{VIX}), \textit{Invesco DB Commodity Index Tracking Fund} (\texttt{DBC}) and Vanguard  Total Stock Market Index Fund (\texttt{VTI}).

As the second dataset (named \texttt{STOCKS} in the text) we have obtained the spot prices of 66 securities from the Nasdaq index, from January 2nd 2003 to December 30th 2022. We have selected this subset of 66 stocks as it was the longest subsection of Nasdaq data with no missing data.
All the data have been obtained from Yahoo Finance.

As the intermediate size dataset (named \texttt{FAMAFRENCH} in the text), we considered the daily returns from 10 industry portfolios starting from January 1st 1973 until November 30th 2023. Data were obtained from the Fama-French data library.\footnote{\url{https://mba.tuck.dartmouth.edu/pages/faculty/ken.french/ftp/10_Industry_Portfolios_daily_CSV.zip}}

Finally, we have run experiments on commodities future prices downloaded from Yahoo finance,\footnote{\url{https://finance.yahoo.com/}} from 2003 to 2023. The dataset named \texttt{COMMODITIES} included 19 commodities from gold to gasoline and soybean.
All the datasets tickers list is provided in the Supplementary Materials Section.
\begin{table}
\centering
\begin{tabular}{lcccc}\hline
\toprule
\textbf{Dataset}     & \textbf{\#assets} & \textbf{Start} & \textbf{End} & \textbf{Coverage} \\
\midrule
\texttt{ETF}         & 4 & 2006-02-06 & 2023-12-29 & $\approx15$ years\\ 
\texttt{FAMAFRENCH}  & 10 & 1973-01-02 & 2023-11-30 & $\approx50$ years\\
\texttt{STOCKS}      & 66 & 2003-01-02 & 2022-12-30 & $\approx20$ years \\
\texttt{COMMODITIES} & 19 & 2003-01-02 & 2023-12-29 & $\approx20$ years \\
\bottomrule
\end{tabular}
\caption{Properties of the datasets used through this work.}
\end{table}

\subsection{Benchmarks}
We have compared the results of our two Hopfield-based models to the traditional convex optimization methods designed to look for portfolios with the least risk.
First and foremost, the Markowitz Mean-Variance portfolio~\citep{markowitz1952} where we have looked at the minimum volatility solution (MVO).
Second, we have selected the excellent Hierarchical Risk Parity method by \citet{deprado2018advances}, indicated as HRP. The entire data without any batching was used for these convex optimisation methods.
To make the comparison fair with another deep learning method, we have replicated the algorithm by \citet{zhang2020deep} using the LSTM network. However, we used only the asset returns as input, instead of concatenating asset prices and returns, to ensure consistency in the input data for all the methods.
Finally, we evaluated the Hopfield Pooling method (\texttt{HOP-POOL}) and the Hopfield Transformer method (\texttt{HOP-TRA}).

\section{Results}
\begin{figure*}[htb]
\centering

\begin{subfigure}[t]{0.49\textwidth}
    \centering
    \texttt{ETF}\\
    \includegraphics[width=0.95\linewidth]{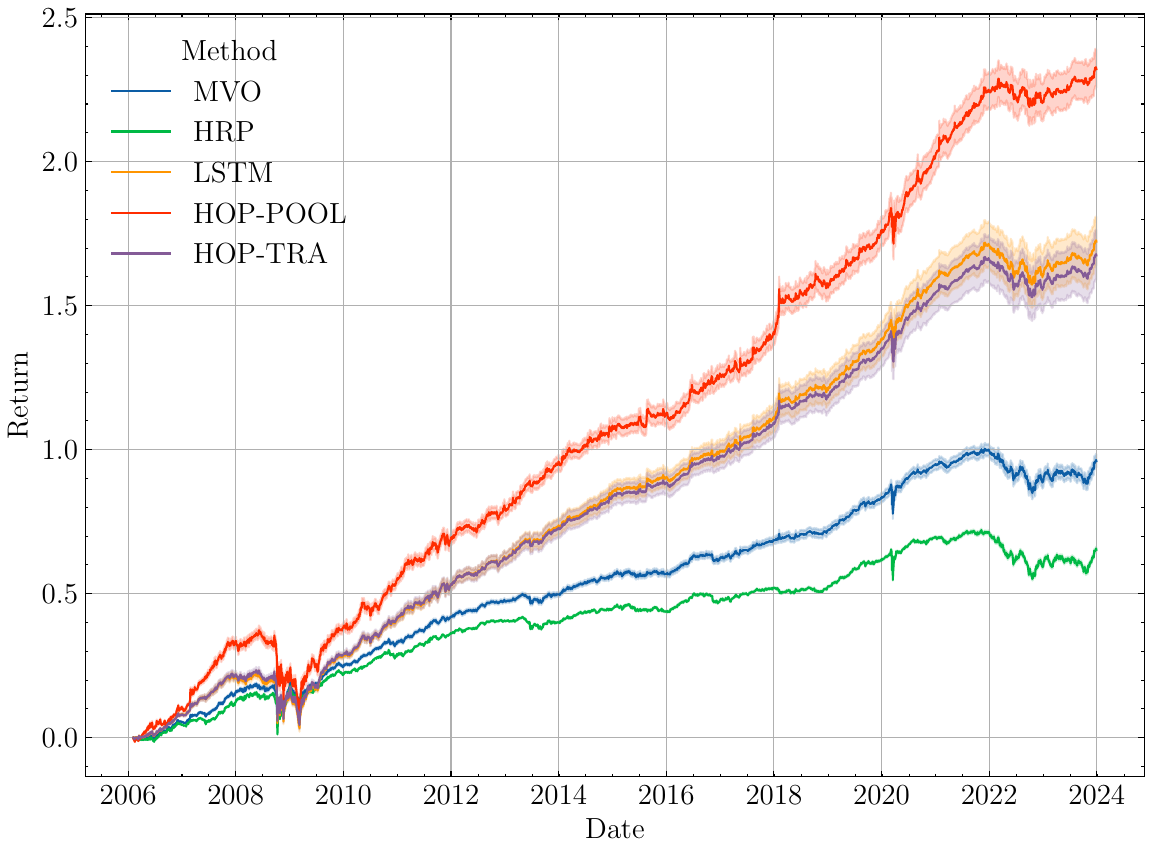}
\end{subfigure}%
\hfill
\begin{subfigure}[t]{0.49\textwidth}
    \centering
    \texttt{FAMAFRENCH}\\
    \includegraphics[width=0.95\linewidth]{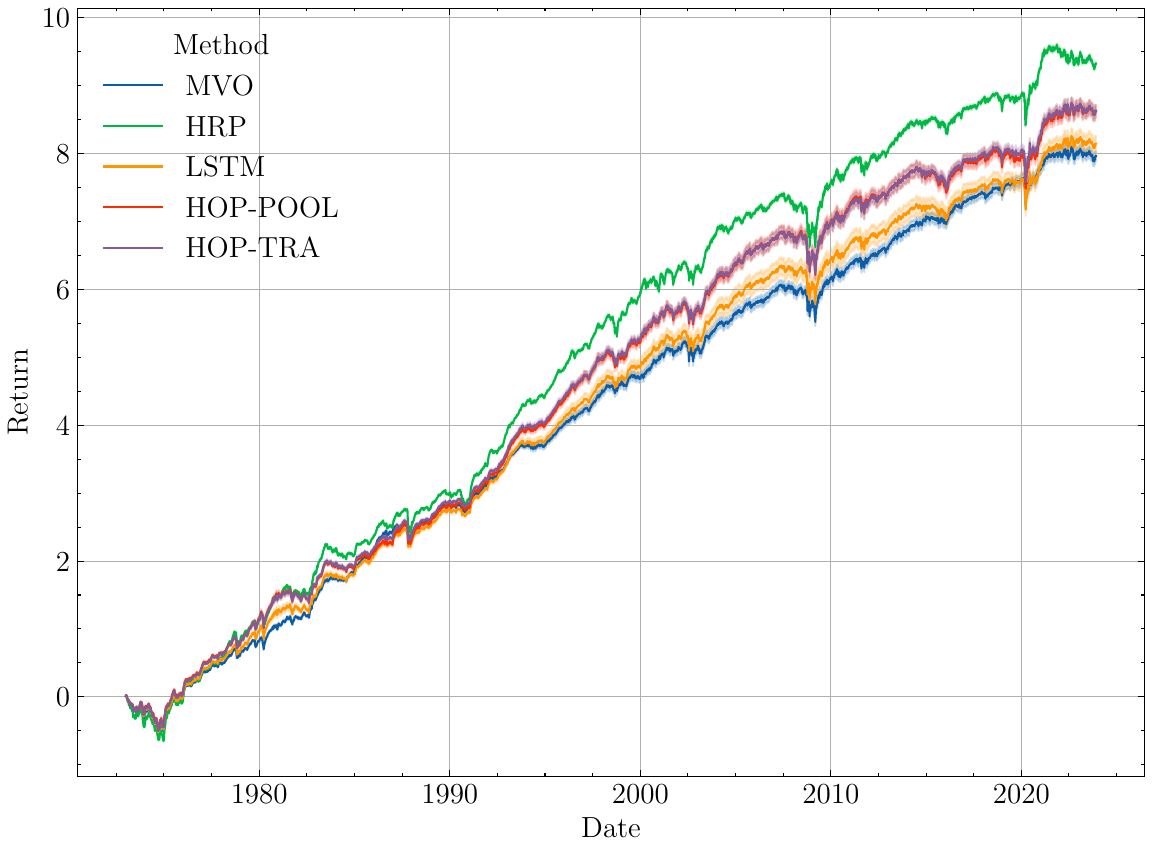}
\end{subfigure}
\\
\begin{subfigure}[t]{0.49\textwidth}
    \centering
    \texttt{STOCKS}\\
    \includegraphics[width=0.95\linewidth]{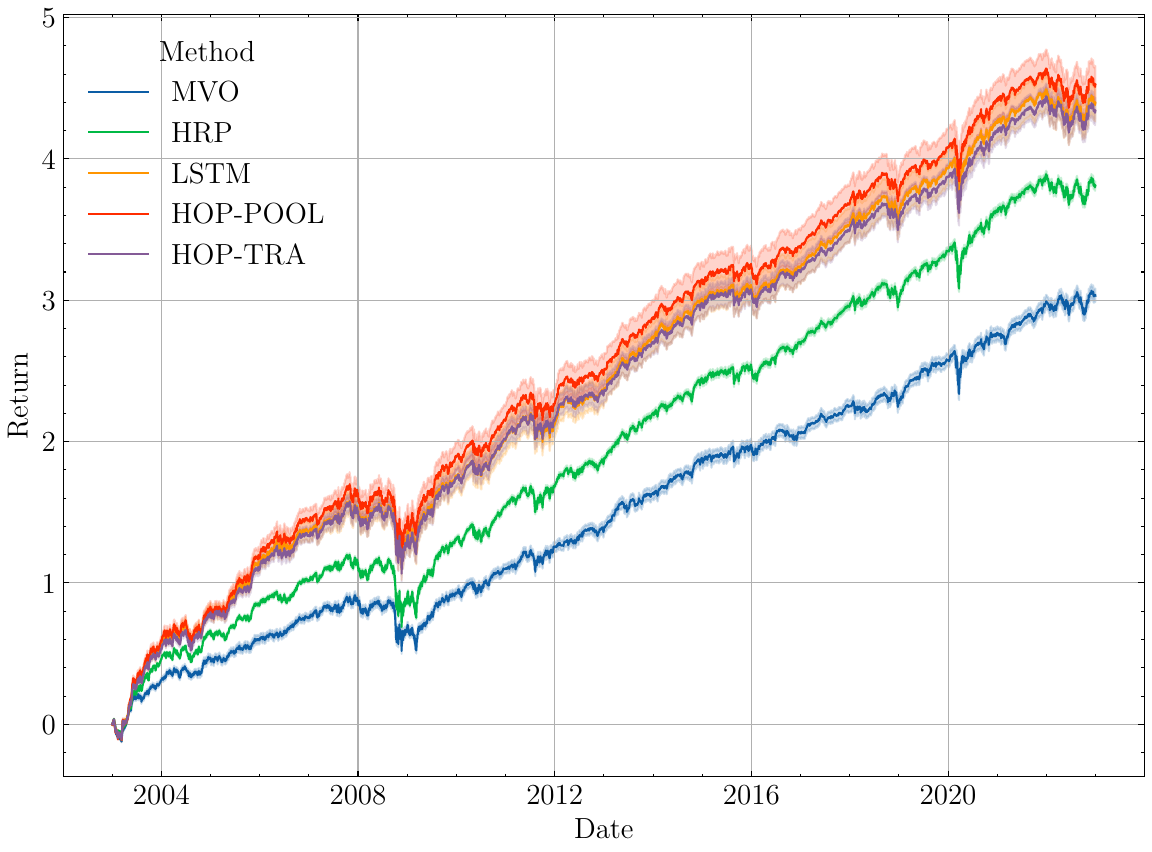}
\end{subfigure}%
\hfill
\begin{subfigure}[t]{0.49\textwidth}
    \centering
    \texttt{COMMODITIES}\\
    \includegraphics[width=0.95\linewidth]{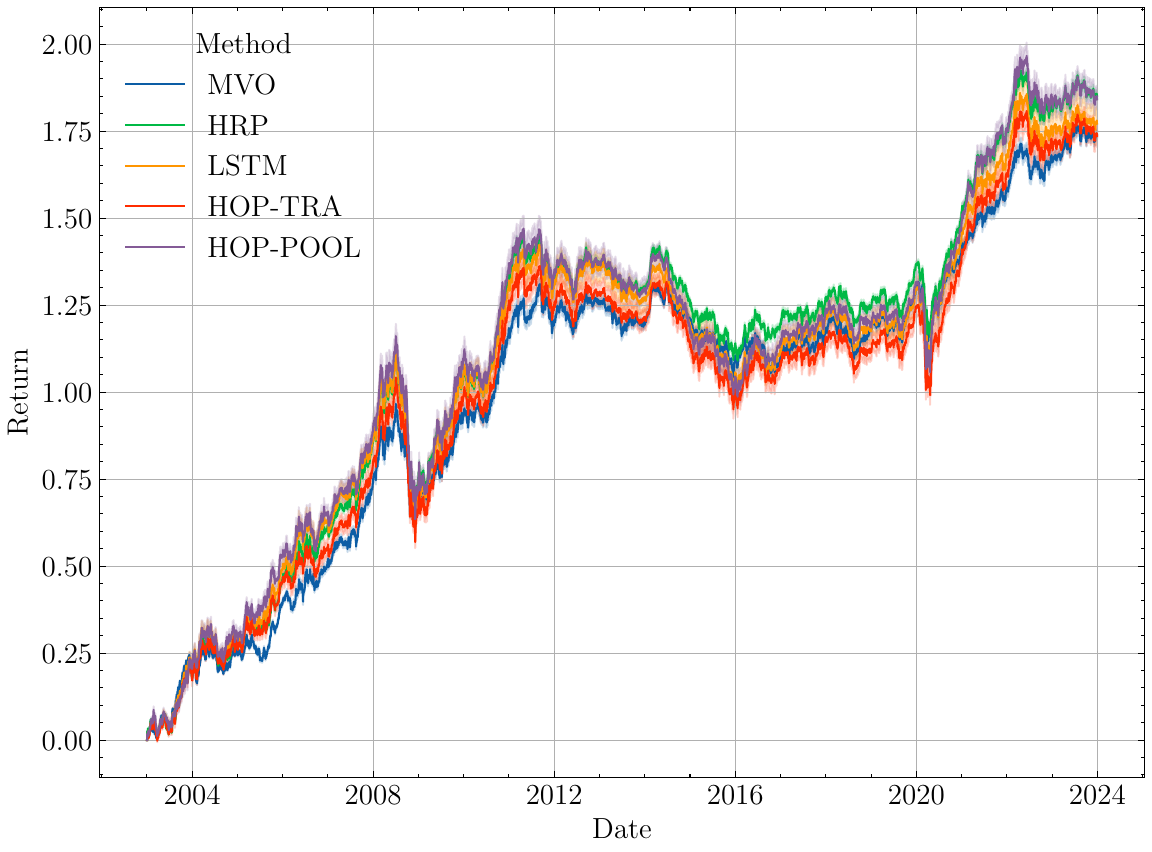}
\end{subfigure}

\caption{Long only portfolios. Cumulative returns over combinatorial purged cross-validation for the four datasets.}
\label{fig:results}
\end{figure*}

We have run several portfolio optimization methods using combinatorial purged cross validation.
Each backtesting path resulted in different test returns' series, possibly with small intersections among the testing paths, which we tried to minimize setting both embargo and purging.
Over all the backtesting paths we have computed the average and standard deviation of several metrics, specifically the annualized average return, annualized Sharpe ratio, annualized Sortino ratio, and average drawdown. 
Moreover, with all the backtesting paths we were able to conduct accurate statistical analyses on portfolio metrics to assess the significance of our claims.
Results are shown in Table~\ref{tab:results} and Figure~\ref{fig:results}.

\begin{figure*}[htb]
\centering

\begin{subfigure}[t]{0.49\textwidth}
    \centering
    \texttt{ETF}\\
    \includegraphics[width=0.95\linewidth]{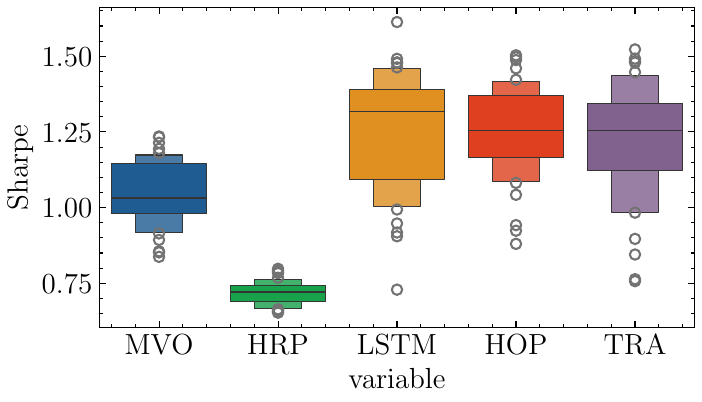}
\end{subfigure}%
\hfill
\begin{subfigure}[t]{0.49\textwidth}
    \centering
    \texttt{FAMAFRENCH}\\
    \includegraphics[width=0.95\linewidth]{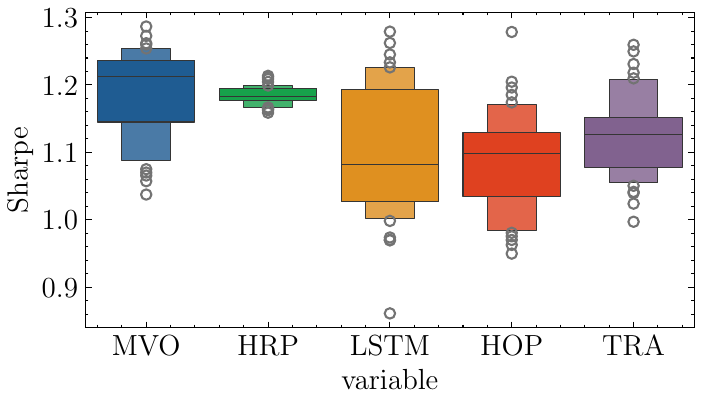}
\end{subfigure}
\\
\begin{subfigure}[t]{0.49\textwidth}
    \centering
    \texttt{STOCKS}\\
    \includegraphics[width=0.95\linewidth]{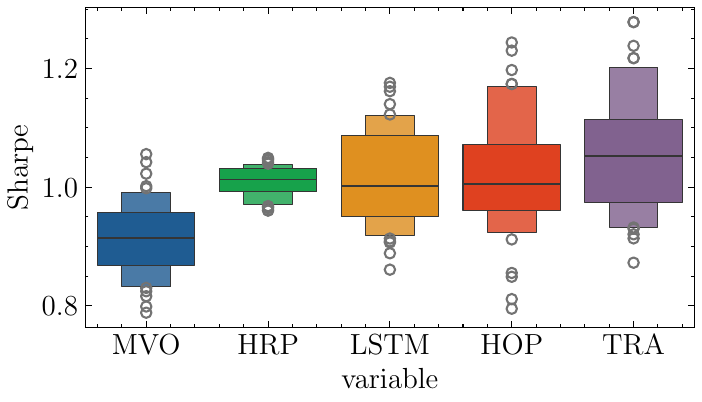}
\end{subfigure}%
\hfill
\begin{subfigure}[t]{0.49\textwidth}
    \centering
    \texttt{COMMODITIES}\\
    \includegraphics[width=0.95\linewidth]{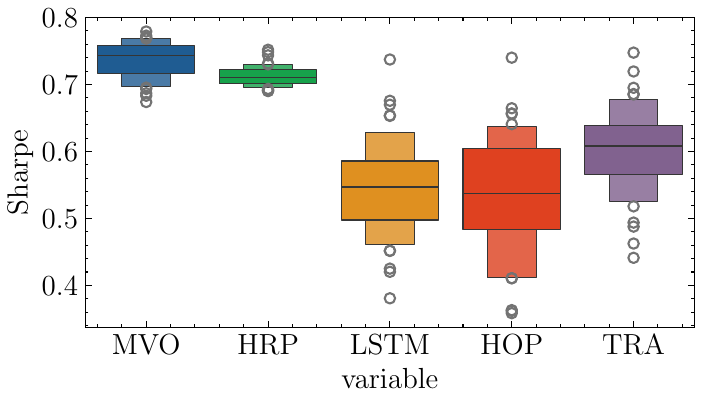}
\end{subfigure}

\caption{Distribution of Sharpe ratios over combinatorial purged cross-validation for the four datasets in the long-only case.} 
\label{fig:results_sharpe}
\end{figure*}

\scriptsize
\begin{table*}[htb]
\begin{tabular}{rcccccc}
\toprule
& \textbf{EW} & \textbf{MVO} & \textbf{HRP} & \textbf{LSTM} & \textbf{HOP-POOL} & \textbf{HOP-TRA} \\
\midrule
\multicolumn{7}{c}{\texttt{ETF}} \\
Mean & $0.227 $ & $0.054 \pm 0.007$ & $0.036 \pm 0.002$ & $0.096 \pm 0.028$ & $\mathbf{0.130 \pm 0.022}$ & $0.094 \pm 0.029$ \\
Sharpe & $0.827 $ & $1.047 \pm 0.112$ & $0.719 \pm 0.041$ & $1.248 \pm 0.205$ & $\mathbf{1.252 \pm 0.163}$ & $1.219 \pm 0.203$ \\
Sortino & $1.404 $ & $1.443 \pm 0.180$ & $0.964 \pm 0.055$ & $1.847 \pm 0.318$ & $\mathbf{1.919 \pm 0.247}$ & $1.799 \pm 0.331$ \\
Avg. DD & $0.070 $ & $\mathbf{0.016 \pm 0.001}$ & $0.019 \pm 0.001$ & $0.020 \pm 0.006$ & $0.026 \pm 0.009$ & $0.020 \pm 0.008$ \\
\multicolumn{7}{c}{\texttt{FAMAFRENCH}} \\
Mean & $0.188 $ & $0.156 \pm 0.008$ & $\mathbf{0.183 \pm 0.003}$ & $0.160 \pm 0.014$ & $0.169 \pm 0.011$ & $0.169 \pm 0.010$ \\
Sharpe & $1.157 $ & $\mathbf{1.190 \pm 0.068}$ & $1.184 \pm 0.014$ & $1.099 \pm 0.098$ & $1.087 \pm 0.077$ & $1.122 \pm 0.064$ \\
Sortino & $1.557 $ & $\mathbf{1.620 \pm 0.085}$ & $1.595 \pm 0.017$ & $1.501 \pm 0.126$ & $1.481 \pm 0.099$ & $1.526 \pm 0.082$ \\
Avg. DD & $0.078 $ & $\mathbf{0.045 \pm 0.003}$ & $0.073 \pm 0.003$ & $0.057 \pm 0.015$ & $0.062 \pm 0.014$ & $0.058 \pm 0.009$ \\
\multicolumn{7}{c}{\texttt{STOCKS}} \\
Mean & $0.218 $ & $0.152 \pm 0.013$ & $0.191 \pm 0.008$ & $0.220 \pm 0.035$ & $\mathbf{0.227 \pm 0.039}$ & $0.217 \pm 0.027$ \\
Sharpe & $1.019 $ & $0.915 \pm 0.067$ & $1.010 \pm 0.027$ & $1.014 \pm 0.087$ & $1.018 \pm 0.109$ & $\mathbf{1.059 \pm 0.107}$ \\
Sortino & $1.407 $ & $1.275 \pm 0.093$ & $1.400 \pm 0.036$ & $1.420 \pm 0.125$ & $1.424 \pm 0.154$ & $\mathbf{1.474 \pm 0.151}$ \\
Avg. DD & $0.049 $ & $\mathbf{0.035 \pm 0.005}$ & $0.039 \pm 0.002$ & $0.050 \pm 0.009$ & $0.053 \pm 0.011$ & $0.045 \pm 0.007$ \\
\multicolumn{7}{c}{\texttt{COMMODITIES}} \\
Mean & $0.098$ & $\mathbf{0.083\pm0.003}$ & $0.088\pm0.003$ & $0.084\pm0.012$ & $0.083\pm0.009$ & $0.088\pm0.008$ \\
Sharpe & $0.658$ & $\mathbf{0.736\pm0.029}$ & $0.713\pm0.016$ & $0.545\pm0.076$ & $0.600\pm0.069$ & $0.536\pm0.096$ \\
Sortino & $0.911$ & $\mathbf{1.009\pm0.042}$ & $0.978\pm0.022$ & $0.755\pm0.105$ & $0.747\pm0.129$ & $0.827\pm0.095$ \\
Avg. DD & $0.155$ & $\mathbf{0.084\pm0.013}$ & $0.115\pm0.014$ & $0.155\pm0.054$ & $0.139\pm0.036$ & $0.181\pm0.075$ \\
\bottomrule
\end{tabular}
\vspace{2mm}
\caption{Results of combinatorial purged cross validation. \texttt{HOP-POOL} and \texttt{HOP-TRA} have very similar performance in terms of Sharpe ratio.}
\label{tab:results}
\end{table*}
\normalsize

Moreover, in order to provide a better assessment of the statistical relevance of our analyses we have run a Tukey HSD test~\citep{tukey1949comparing}, a method for comparing the means of multiple groups that highlights which pairs are significantly different from each other.
We have run the Tukey test independently over all the four datasets presented in this work. 
In Table~\ref{tab:tukey_cld} we indicate with Compact Letters Display notation~\citep{piepho2004cldalgorithm} the similarity groups of methods' Sharpe ratios  among the different datasets.

\begin{table}[htb]
\centering
\begin{tabular}{rcccc}
\toprule
 & \texttt{ETF} & \texttt{STOCKS} & \texttt{FAMAFRENCH} & \texttt{COMMODITIES}\\
\midrule
\texttt{MVO} & c & b & b & b \\
\texttt{HRP} & b & a & b & b \\
\texttt{LSTM} & a & a & a & c \\
\texttt{HOP-POOL} & a & a & a & c \\
\texttt{HOP-TRA} & a & a & a & a \\
\bottomrule
\end{tabular}
\vspace{2mm}
\caption{Compact Letters Display notation for the Sharpe ratio after Tukey HSD test on long-only results. Each column indicates to what statistically similar group the specific method belongs. The full Tukey pairwise test is shown in the Supplementary Materials Section.}
\label{tab:tukey_cld}
\end{table}
 
To summarize the findings from the statistical analyses of Table~\ref{tab:tukey_cld} we can conclude that in the \texttt{ETF}, \texttt{STOCKS} and \texttt{FAMAFRENCH} datasets the three deep learning methods perform similarly. 
However the Hopfield Encoder performs better than both LSTM and Hopfield Pooling in the \texttt{COMMODITIES} dataset.

Additionally we visually explored the distribution of backtesting Sharpe ratios through a Box plot in Figure~\ref{fig:results_sharpe}.

The results, taken over all the four datasets are very encouraging, although some weakness of the proposed methods is evidenced. 
On the smallest dataset (\texttt{ETF}), we reach high Sharpe ratio consistently, with the statistical tests indicating that our proposed method is not statistically different from the \texttt{LSTM}, but presents statistically relevant higher Sharpe ratios with respect to the two other traditional methods \texttt{MVO} and \texttt{HRP}.
The superiority of the end-to-end deep learning approaches (\texttt{LSTM}, \texttt{HOP-POOL}, \texttt{HOP-TRA}) are also clear in the \textsf{STOCKS} dataset, again with statistically significant higher Sharpe ratios between the deep learning methods against the traditional ones. 
In this case however all the deep learning methods perform similarly.

Another kind of results are highlighted for the \texttt{COMMODITIES} and \texttt{FAMAFRENCH} datasets. 
Here all the deep learning methods fail in comparison to traditional approaches. 
We believe that this result stems from a poor generalization effect in return series where multiple regimes are present.
Indeed, differently from the \texttt{STOCKS} and \texttt{ETF} datasets where a growth trend is evident, the \texttt{COMMODITIES} and \texttt{FAMAFRENCH} are much more variable.
Hence, taking an average of almost two years of predicted weights over highly variable regimes may hinder the beneficial effects observed with other methods.

\paragraph{Equally weighted portfolio.} 
In the aforementioned analysis, we only reported the cross-validation metrics for optimization based methods with their uncertainty computed as the standard deviation of each metric over the backtesting paths.
We omitted the metrics uncertainty for the equally weighted portfolio as each metric had the same value over all backtesting paths, hence zero standard deviation.
We also note that the equally weighted portfolio allocation has serious problems in the datasets analyzed. First, in the \texttt{ETF} dataset, allocating 25\% to the VIX index was deemed unfavorable due to its association with excessive volatility, despite yielding a commendable final total return.
In the \texttt{STOCKS} dataset, comprising 66 assets, adhering to an equally weighted portfolio approach would lead to impractical transaction costs during rigorous back-testing procedures.
The same applies to the \texttt{FAMAFRENCH} and \texttt{COMMODITIES} datasets.
Nonetheless we report the equally weighted backtesting results in a separate chart for completeness in the Supplementary Materials section.

\section{Discussion}
In this section we discuss some of the observations resulting from the experiments, together with possible interpretations and comments.
\paragraph{Hopfield pooling networks work like LSTMs.}
Our work shows that modern Hopfield networks can reliably be used also in the time series domain, specifically in non-standard tasks like the one of portfolio optimization.
What we have found is that a LSTM network can be replaced with an Hopfield Pooling network, even without the need of a Time2Vec embedding. 
Our experiments show that most of the times the results from the \texttt{LSTM} and \texttt{HOP-POOL} models are similar, but training \texttt{HOP-POOL} is faster, given the lower number of parameters.

\paragraph{A Hopfield network can replace the multihead self attention layer.}
In the proposed Hopfield Encoder architecture, we have observed that the Hopfield layer can replace the classical multi-head self-attention layer used in the Transformer block.
In our experiments we have not run any hyper-optimization on the multitude of parameters of the encoder architecture, like number of layers, number of heads within each Hopfield layer or internal embedding dimension, instead we have chosen the same parameters as in the original implementation by \citet{ramsauer2020hopfield}.
However, as shown in the previous section, results are encouraging and while we see slightly lower Sharpe ratio, this happens at both better average and maximum drawdown. 
Moreover in the hardest dataset (\texttt{COMMODITIES}) we have statistically higher Sharpe ratio for the Hopfield Encoder with respect to both LSTM and Hopfield Pooling.
In future, we may perform a large hyper-parameters search in order to evaluate how the results are influenced by an optimal selection of parameters. However, we are confident that an accurate hyperparameters' search could yield even better results in downstream tasks.

\paragraph{Combinatorial purged cross-validation and statistical analysis.}
This research marks one of the few attempts to employ combinatorial purged cross-validation extensively in testing various backtesting paths, particularly with the application of deep-learning models. 
Using this method offers the benefit of providing a more reliable assessment of a portfolio optimization technique across numerous historical scenarios and helps reduce the likelihood of false discoveries. 
Consequently, we strongly advise against using the standard walk-forward cross-validation in finance, as it tends to test a single path with a pronounced over-fitting effect. 
This approach may compromise the accuracy and robustness of the results. 
Statistically accurate analyses on an even larger number of backtesting paths can provide a better view on the future properties of all allocation methods.

\section{Conclusions}
In the context of portfolio optimization, our research aims to harness the capabilities of deep learning  to enhance the decision-making process of asset allocation.

The application of deep learning in this context aligns with the broader trend in the financial industry to integrate advanced technologies for more sophisticated portfolio strategies.
Our exploration of the applicability of modern Hopfield networks in the realm of portfolio optimization signifies a departure from conventional approaches.
This renewed interest is driven by the potential advantages that Hopfield networks may offer in handling the inherent complexities of optimization tasks, providing an alternative avenue for practitioners to explore.
The two methods proposed in this work - Hopfield Pooling and Hopfield Encoder, demonstrates better or comparable performance against other traditional and deep-learning based methods in most of the tasks analyzed.

In this study, our focus has been solely on analyzing returns data. 
However, the deep architectures we have introduced can be effectively applied to model unconventional data sources, provided they are appropriately processed. 
Examples of alternative data sources include Environmental, Social, and Governance (ESG) factors or macroeconomic sentiment indicators.
Traditional methods based on convex optimization encounter challenges when incorporating such diverse data sources.
In contrast, deep models only necessitate minor architectural adjustments to accommodate them.

It is essential to note that the methods outlined in this work do not claim to be a definitive solution for portfolio optimization tasks; rather, they represent an additional tool to consider in addressing complex challenges posed to asset managers.
Indeed, the primary value proposition of our work lies in the exploration of how deep-learning based methods, specifically Hopfield networks, can augment traditional decision-making processes, offering an alternative perspective.

As a future research we would like to include the ability to add specific constraints into the model optimization. We believe that the ability to handle complex constraints like cardinality constraints, or minimum risk budget could help the quantitative analyst to better align the raw algorithm results with the markets views imposed by the portfolio managers.
In this study, our primary focus was on cross-validation metrics. However, we did not delve into exploring how fine-tuning hyper-parameters could potentially improve allocations, nor did we examine the impact of assets turnover and management fees on the final strategy.
These areas warrant further investigation, which we leave to subsequent studies.

\section*{Broader Impact}
This study highlights new areas where modern Hopfield networks can be beneficial, particularly in finance, potentially opening to the possibility of including new kind of data apart from the traditional asset returns, such as NLP embeddings or generally temporal trends and market sentiments.
We demonstrate how these networks can enhance portfolio optimization, providing a valuable tool for financial practitioners.

Our findings show that Hopfield networks can improve the efficiency and robustness of portfolio management. They represent a powerful addition to the existing optimization techniques, helping professionals handle complex financial data more effectively.
Overall, this research expands the applications of Hopfield networks and offers practical insights for their use in financial optimization.

\bibliographystyle{ACM-Reference-Format}
\bibliography{biblio}

\end{document}